# A Frequency-Velocity CNN for Developing Near-Surface 2D Vs Images from Linear-Array, Active-Source Wavefield Measurements

Aser Abbas [a, *], Joseph P. Vantassel [b], Brady R. Cox [a], Krishna Kumar [c], Jodie Crocker [c]

[a] Utah State University, Department of Civil and Environmental Engineering, Logan, UT, USA, 84322.

[b] Virginia Tech, Department of Civil and Environmental Engineering, Blacksburg, VA, USA, 24061.

[c] The University of Texas at Austin, Department of Civil, Architectural, and Environmental Engineering, Austin, TX, USA, 78712.

## Abstract

This paper presents a frequency-velocity convolutional neural network (CNN) for rapid, non-invasive 2D shear wave velocity ($V_S$) imaging of near-surface geo-materials. Operating in the frequency-velocity domain allows for significant flexibility in the linear-array, active-source experimental testing configurations used for generating the CNN input, which are normalized dispersion images. Unlike wavefield images, normalized dispersion images are relatively insensitive to the experimental testing configuration, accommodating various source types, source offsets, numbers of receivers, and receiver spacings. We demonstrate the effectiveness of the frequency-velocity CNN by applying it to a classic near-surface geophysics problem, namely, imaging a two-layer, undulating, soil-over-bedrock interface. This problem was recently investigated in our group by developing a time-distance CNN, which showed great promise but lacked flexibility in utilizing different field-testing configurations. Herein, the new frequency-velocity CNN is shown to have comparable accuracy to the time-distance CNN while providing greater flexibility to handle varied field applications. The frequency-velocity CNN was trained, validated, and tested using 100,000 synthetic near-surface models. The ability of the proposed frequency-velocity CNN to generalize across various acquisition configurations is first tested using synthetic near-surface models with different acquisition configurations from that of the training set, and then applied to experimental field data collected at the Hornsby Bend site in Austin, Texas, USA. When fully developed for a wider range of geological conditions, the proposed CNN may ultimately be used as a rapid, end-to-end alternative for current pseudo-2D surface wave imaging techniques or to develop starting models for full waveform inversion.

*Corresponding author.
E-mail: aser.abbas@usu.edu (A. Abbas)

## 1. Introduction

Non-invasive subsurface imaging techniques based on stress wave propagation have gained increased interest over the past few decades due to their significant cost savings over traditional invasive site characterization methods and their potential to cover large areas. The current study proposes a frequency-velocity domain, deep-learning technique for rapid, non-invasive 2D shear wave velocity ($V_S$) imaging of near-surface geo-materials. The ultimate vision of this type of imaging approach is that when fully developed for a wide range of geological conditions, it may be used as a cost-effective and accurate alternative to current pseudo-2D multichannel analysis of surface waves (MASW) (e.g., Park 2005) imaging, or in developing 2D starting models for more rigorous imaging methods like full waveform inversion (FWI) (Tarantola, 1984; Mora,1987). The information provided below illustrates the need for new approaches to near-surface site characterization, presents background information on previous work related to deep-learning subsurface imaging, and explains why a frequency-velocity domain approach is desirable for field application flexibility.

At depths of greatest interest to geotechnical engineering (less than ~ 30 m), surface waves dominate the energy of the elastic wavefield (Miller and Pursey, 1955). As a result, surface wave methods are the most common techniques for developing 1D Vs profiles for near-surface application (e.g., Stokoe et al., 1994; Park et al., 1999; Foti, 2000; Louie, 2001; Okada, 2003; Tokimatsu et al., 1992). Surface wave methods work by exploiting the dispersive properties of surface waves in vertically heterogeneous media to develop 1D Vs profiles through the solution of an inverse problem. Solving the inverse problem involves assuming a 1D model with elastic properties (Foti et al., 2014; 2018) and iteratively solving a theoretical wave propagation problem (i.e., the forward problem) until the theoretical dispersion curves from the assumed model match the dispersion data extracted from experimental measurements of surface waves phase velocity. The surface wave inverse problem, commonly referred to as surface wave inversion, has been explored extensively in the literature and is known to be particularly challenging due to it being ill-posed and without a unique solution (Vantassel & Cox 2021a, b; Cox & Teague 2016; Foti et al., 2014; 2018). Despite these challenges, surface wave methods have been applied widely in practice and are commonly used to develop 1D, and even pseudo-2D, subsurface models, for example with methods such as 2D MASW (Park 2005; Ivanov et al., 2006). It is important to note, however, that these are not true 2D models due to underlying 1D assumptions in the numerical solution of the dispersion data forward problem used during inversion. Thus, the process of spatially interpolating between numerous 1D Vs profiles collected along a linear array produces a pseudo-2D subsurface image rather than a true 2D subsurface image. Presently, the only linear-array, active-source subsurface imaging method capable of producing a true 2D subsurface image is FWI.

While FWI can produce a true 2D model, it is less commonly used than surface wave methods for near-surface imaging due to its more complex and time-consuming field acquisition and data processing requirements. However, it is a more promising approach for recovering true 2D and 3D subsurface images, as it utilizes the entirety of the seismic wavefield (rather than only the surface wave dispersion in surface wave methods). FWI can be described as a data-fitting procedure that seeks to minimize the misfit between the experimentally acquired seismic waveforms and the synthetic wavefield obtained by solving a wave propagation simulation through a candidate model. The FWI optimization process can be performed using either a global or a local search algorithm. Even though numerical methods for modeling the propagation of elastic waves through 2D and 3D earth models exist (e.g., finite-difference, spectral-element), they are computationally expensive, making global search methods, which are already computation demanding, uncommon for FWI (Virieux and Operto 2009). Local search methods (e.g., Pratt et al., 1998; Pratt 1999; Nocedal and Wright 2006) are less computationally expensive, as they begin with a predefined starting model and iteratively refine that model until the misfit between the recorded seismic waveforms and the calculated wavefield becomes sufficiently small. As a result, they require solving fewer forward problems than their global counterparts, making them computationally less expensive. However, if the starting model is not sufficiently similar to the true subsurface model, these methods are likely to be trapped in a local minimum, or saddle point, that prevents them from converging to the true solution (Monteiller et al., 2015; Smith et al., 2019; Feng et al., 2021; Vantassel and Cox 2022). Given the sensitivity of FWI results to the starting model (Shah et al., 2012; Vantassel et al., 2022a), rapid and accurate ways of generating 2D and 3D starting models are needed to more fully take advantage of FWI in engineering practice.

There has been growing interest in the past few years in using deep-learning methods to either enhance or completely replace FWI. An extensive review on integrating these deep-learning methods in various parts of the FWI can be found in (Alder et al., 2021). Several end-to-end techniques, which aim to retrieve subsurface models directly from seismic wavefield data are also available in the literature (e.g., Araya-polo et al., 2018; Mosser et al., 2018a, b; Mao et al., 2019; Yang and Ma, 2019; Li et al., 2020). However, most previous works have either targeted recovering crustal-scale subsurface velocity models, as indicated by Vantassel et al. (2022a), or/and suffered from a weak generalization ability, preventing them from being used for a wide variety of field applications, as noted by Feng et al. (2021) and Liu et al. (2020). Additionally, a significant portion of the deep learning seismic imaging literature has focused on developing 2D velocity models using the acoustic approximation (Araya-polo et al., 2018; Mosser et al., 2018a, b; Mao et al., 2019; Yang and Ma, 2019; Li et al., 2020), which only models the propagation of compression wave velocities ($V_P$) and makes them poorly suited for near-surface applications. In fully saturated near-surface soil deposits, $V_P$ is mainly controlled by the compressibility of the water and is much faster than the $V_P$ of dry materials (Foti et al., 2014). Hence, $V_P$ is less revealing of actual subsurface soil properties when the materials are relatively soft and saturated. On the other hand, the small strain shear modulus calculated from $V_S$ represents that of the soil skeleton only and is independent of the ground saturation (Aziman et al., 2016). Due to the interest in retrieving Vs for engineering site characterization and the predominance of surface waves in actively-generated wavefields, deep learning approaches based only on acoustic wave propagation are not applicable for near-surface site characterization. To the authors' knowledge, the sole deep-learning, end-to-end, 2D imaging technique for near-surface geotechnical engineering purposes was proposed by Vantassel et al. (2022a).

Vantassel et al. (2022a) demonstrated the ability of deep-learning methods in utilizing complicated wavefields comprised of surface and body waves to image the near-surface. They designed a CNN that could be used to generate 2D Vs images for subsurface profiles consisting of soil over undulating rock. Their CNN could predict a 24-m deep and 60-m wide $V_S$ image directly from waveforms recorded by 24 receivers at 2-m spacing, which is a common configuration for active-source, linear-array imaging techniques for FWI. They used 100,000 synthetic soil-over-rock models in training the CNN and tested it on an additional 20,000 synthetic models. Their CNN showed great promise for developing starting models for near-surface FWI and, in some cases, yielded 2D subsurface models that could not be improved upon by local search FWI. However, the authors acknowledged that their approach could not generalize beyond the data acquisition configurations selected during CNN training (e.g., source type, source location, number of receivers, and receiver spacing).

The present work aims to show that developing a CNN with a frequency-velocity domain input image can yield comparable accuracy to the time-distance domain input approach proposed by Vantassel et al. (2022a), while providing the flexibility necessary to generalize for a broader range of field applications. We demonstrate that once trained for an appropriate set of geological conditions, the proposed frequency-velocity CNN approach can be used to instantly generate a 2D subsurface Vs image directly from a normalized dispersion image obtained from linear-array, active-source wavefield measurements. Normalized dispersion images are shown to be relatively insensitive to the experimental testing configuration and can be easily generated due to their wide use in surface wave testing. The effectiveness of our frequency-velocity CNN is demonstrated by applying it to a classic near-surface geophysics problem; namely, imaging a two-layer, undulating, soil-over-bedrock interface. A total of 100,000 models were developed to train, validate, and test the frequency-velocity CNN. The ability of the proposed frequency-velocity CNN to generalize across various acquisition configurations is first tested using synthetic near-surface models using different acquisition configurations from that of the training set, and then applied to experimental data collected at the Hornsby Bend site in Austin, Texas, USA. We also compare the performance of the frequency-velocity CNN with the time-distance CNN for different near-surface models.

## 2. Overview of the frequency-velocity CNN

The 2D Vs imaging approach proposed herein builds on the work of Vantassel et al. (2022a) by developing a frequency-velocity CNN that is accommodating of different linear-array, active-source experimental testing configurations. A schematic illustrating the similarities and differences between the time-distance CNN (left) and the frequency-velocity CNN (right) required for subsurface Vs imaging is illustrated in Figure 1. The time-distance CNN proposed by Vantassel et al. (2022a) receives an input seismic wavefield recorded at specific receiver locations relative to a single source type and location and predicts a 2D $V_S$ image. The proposed frequency-velocity CNN generalizes beyond specific receiver locations, source types and source locations by using a frequency-dependent normalized dispersion image as its input for predicting a 2D sub-surface $V_S$-image. Unlike the time-distance CNN, which would require additional training and tuning to handle multiple source types and receiver spacings, the frequency-velocity-based CNN is relatively insensitive to the experimental testing configuration and, therefore, saves the network from needing to learn that additional complexity. In both cases, time-distance and frequency-velocity, developing and training the CNN takes a significant amount of time and effort, but after it has been trained it can be used to instantaneously produce a $V_S$ image from the input.

To demonstrate the key difference between training a CNN using a time-distance input (i.e., seismic wavefield) versus a frequency-velocity input (i.e., dispersion image), several synthetic seismic wavefields acquired with different testing configurations on the same subsurface model are shown in Figure 2 along with their associated dispersion images. Figure 2a depicts a synthetic soil-over-rock $V_S$ image. At its surface, 48 receivers with 1-m spacing and two source locations at 5 m and 20 m to the left of the first receiver are shown. Figures 2b through 2i show the seismic wavefields resulting from several different source and receiver configurations and their associated dispersion images. For example, Figures 2b and 2c show the wavefield sampled by 48 receivers at a 1-m receiver spacing and its corresponding dispersion image, respectively, due to a 30-Hz Ricker wavelet source (Figure 3a) at 5-m distance from the first receiver. This will be referred to as the base configuration. The experimental configuration used to obtain the seismic wavefield and dispersion image illustrated in Figures 2d and 2e, respectively, differs from the base configuration in that the source is excited at 20 m from the first receiver. While the wavefield image in Figure 2d is clearly different from the wavefield image of the base configuration (Figure 2b) due to the increased travel time associated with a greater source offset, the dispersion images from both configurations (Figure 2c and 2e) are similar. The experimental configuration used to obtain the wavefield and dispersion image illustrated in Figures 2f and 2g, respectively, differ from the base configuration in that the 30-Hz Ricker wavelet excited 5 m from the first receiver is now recorded by only 24 receivers at a 2-m spacing (i.e., using half the number of receivers at two-times the spacing). Once again, the wavefield image for the alternate testing configuration (Figure 2f) is clearly different from the base configuration (Figure 2b) due to wavefield sampling at one-half the spatial resolution, but the dispersion images (Figures 2c and 2g) are very similar. Figure 2h shows a wavefield sampled by 48 receivers at a 1-m receiver spacing from a source located 5 m from the first receiver (similar to the base configuration), however, the source function is now a 12-second-long linear chirp from 3-Hz to 80-Hz (Figure 3c). Once again, while this wavefield is drastically different from the base configuration wavefield (Figure 2b), its dispersion image is visually identical to the others.

To quantitatively illustrate the good agreement between the dispersion images across different experimental testing configurations, the mean structural similarity index (MSSIM) proposed by Wang et al. (2004) is used. In this case, the MSSIM is used to compare the similarity between the base configuration dispersion image (Figure 2c) and the dispersion images obtained using the varied testing configurations (refer to Figures 2e, 2g, and 2i). The value of the MSSIM index between two images can range between 0 and 1, where a value of 0 indicates no structural similarity, while a value of 1 means perfect structural similarity. More information about MSSIM is provided in section 5.1. In the meantime, the high values of MSSIM for the different testing configurations (0.81 – 1.0) echo the qualitative observations made previously that the four dispersion images are very similar to one another. This insensitivity of dispersion images to acquisition configurations of sources and receivers grants the frequency-velocity CNN approach the flexibility needed to predict on diverse testing configurations independent of the training dataset on which the CNN was trained, which is imperative for field applications.

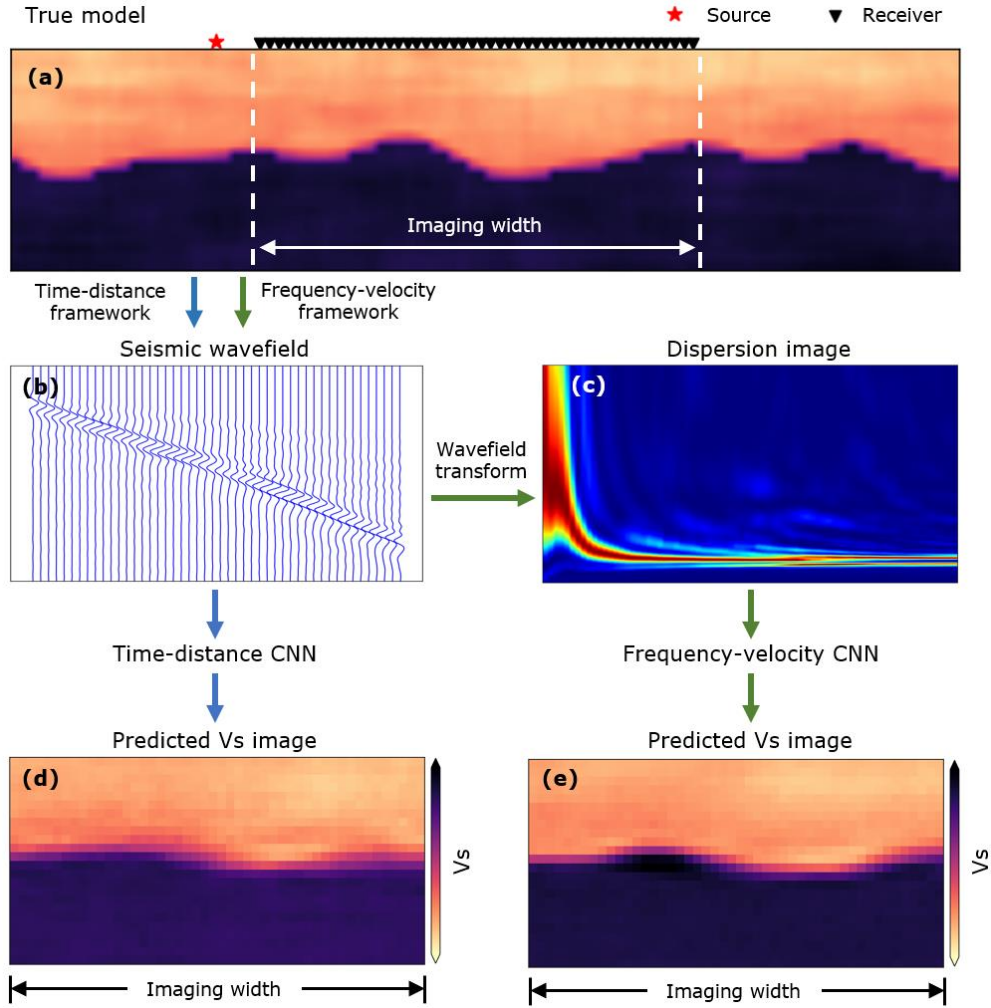

**Fig. 1.** Time-distance CNN framework proposed by Vantassel et al. (2022a), which follows the blue arrow's path (left), and the new frequency-velocity CNN framework, which follows the green arrow's path (right).

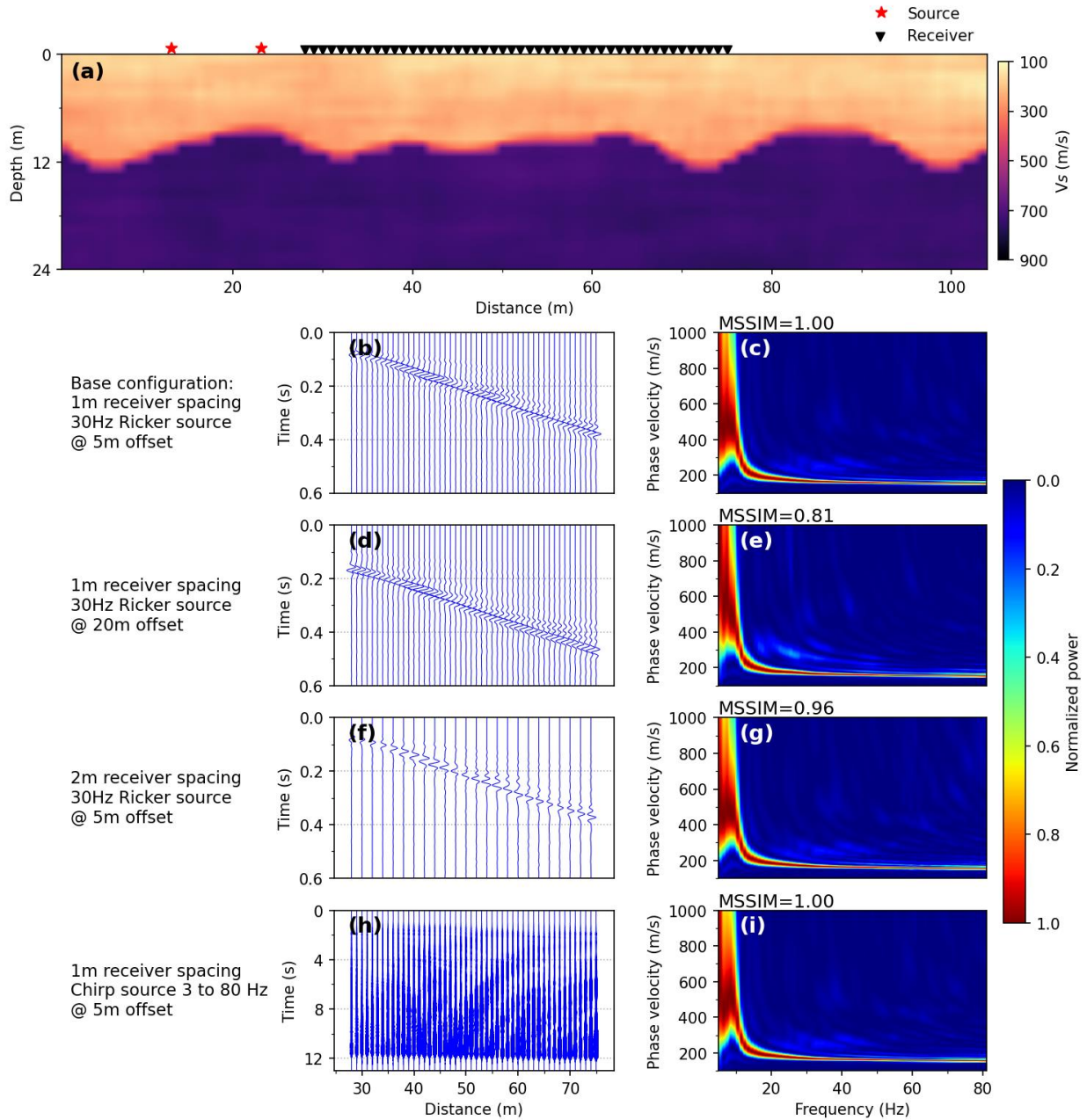

**Fig. 2.** (a) a 104-m wide and 24-m deep soil-over-rock 2D model with a 47-m array of receivers at 1-m spacing and two source locations at 5 and 20 m off the end of the array. (b) the seismic wavefield recorded by 48 receivers from a 30-Hz Ricker source at the 5 m source location and (c) its associated dispersion image. (d) the seismic wavefield recorded by 48 receivers from a 30-Hz Ricker source at the 20 m source location and (e) its associated dispersion image. (f) the seismic wavefield recorded by 24 receivers from a 30-Hz Ricker source at the 5 m source location and (g) its associated dispersion image. (h) the seismic wavefield recorded by 48 receivers from a 3-Hz to 80-Hz chirp/sweep over 12-seconds at the 5 m source location and (i) its associated dispersion image. The mean structural similarity index (MSSIM) of each dispersion image relative to the base case (i.e., panel c) is presented above each panel.

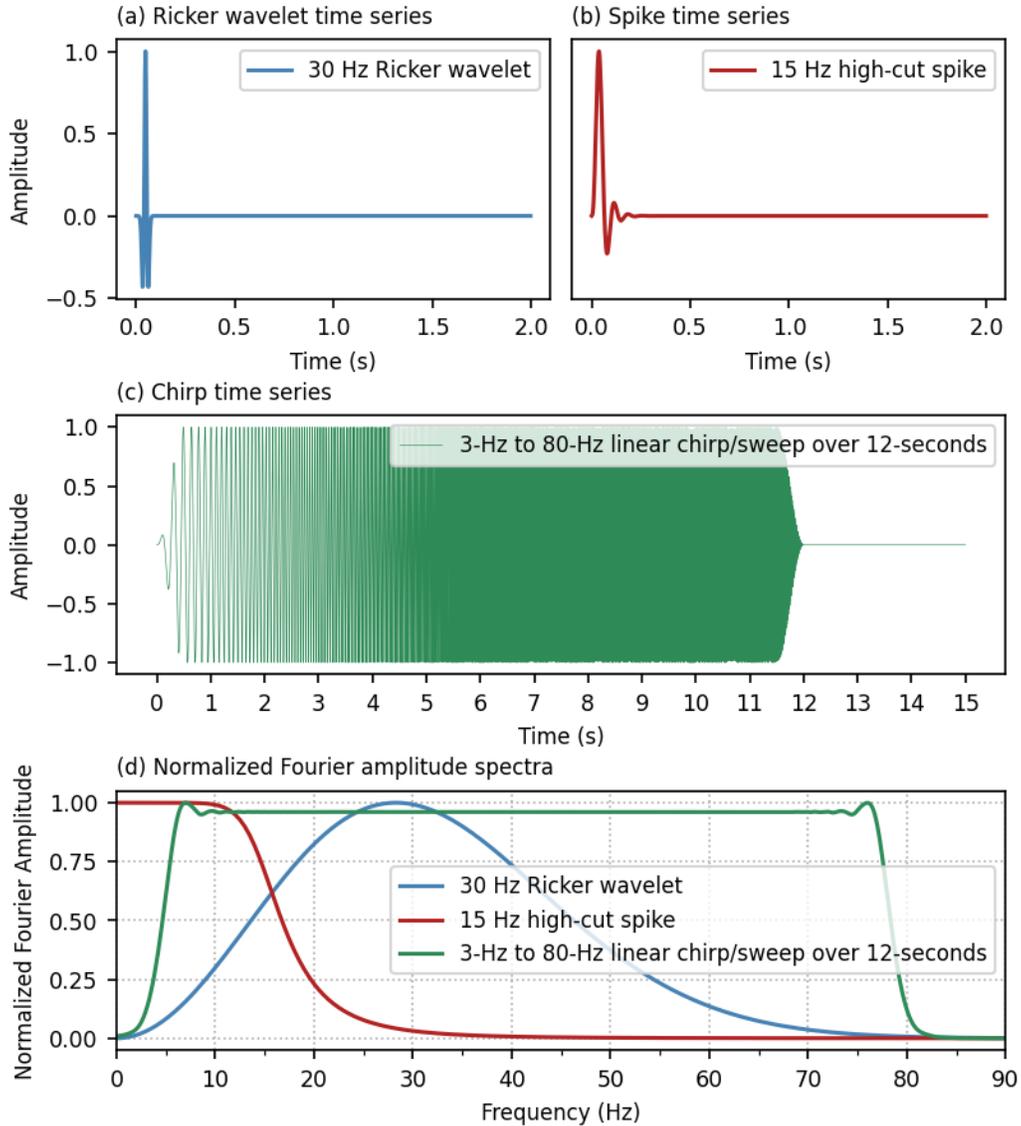

**Fig. 3.** Source functions used in this study and their associated normalized Fourier amplitude spectra. (a) the time series of a 30-Hz Ricker wavelet. (b) the time series of a 15-Hz high-cut filtered spike wavelet. (c) the time series of a 3-Hz to 80-Hz linear chirp/sweep over 12-seconds. (d) the normalized Fourier amplitude spectra of each source function.

The present study is structured as follows. We start by describing the synthetic soil models developed for this study. Next, the details of the wave propagation simulations and the post-processing of the results used in constructing the dataset for this study are presented. We then outline the architectures of the time-distance and frequency-velocity CNNs and evaluate their relative performances on the developed dataset. Subsequently, we demonstrate the advantages of operating in the frequency-velocity domain in terms of flexibility and generalization across different testing configurations that are not present in the training set relative to the time-distance CNN. Lastly, we validate the potential of the frequency-velocity CNN for handling experimental data based on comparisons with field observations at the Hornsby Bend site in Austin, Texas, USA.

## 3. Development of synthetic near-surface models

The synthetic near-surface models developed for this study used a slightly modified version of the framework developed by Vantassel et al. (2022a) for generating realistic soil-over-undulating bedrock subsurface profiles. A brief description of the models with changes relative to those implemented by Vantassel et al. (2022a) is highlighted in this section. A total of 100,000 synthetic near-surface models representative of soil overlying irregular bedrock were developed to train, validate, and test the CNN. A 104-m wide and 24-m deep domain was utilized in the present work. This domain is larger than the 60-m by 24-m domain used by Vantassel et al. (2022a) to accommodate the different receiver and source configurations required for testing the abilities of the proposed frequency-velocity approach to generalize across different field acquisition setups.

Similar to Vantassel et al. (2022a), the base Vs models were constructed by first assuming the vertical variation of Vs in the overlying soil layer (i.e., the upper part of the model) followed the approximate relationship between Vs and mean effective confining pressure for dense granular soils proposed by Menq (2003). To avoid unrealistically low velocities, the Vs relationship was truncated near the ground surface (i.e., at low mean effective stresses) to ensure no Vs less than 200 m/s. To model a broader range of realistic dense soil-velocities, the relationship was scaled up and down by a random variable, the soil velocity factor, between 0.9 and 1.1. The average interface boundary between the stiff soil layer and bedrock ranged between 5 m and 20 m in depth. Several different soil-to-bedrock interface conditions were simulated; namely, 30% of the models had highly undulating interfaces (e.g., Figure 4a through 4f), 60% had slightly undulating interfaces (e.g., Figure 4g through 4r), and 10% of the models had linear soil-rock interfaces (e.g., Figure 4s and 4t). These percentages of models with different bedrock undulation conditions were preserved throughout the CNN training, validation, testing, and generalization evaluation stages. Three overlapping spatial undulation frequencies were used to control the interface undulation intensity. The range of spatial frequencies used for the highly undulating interfaces was between $1/5$ m$^{-1}$ and $1/60$ m$^{-1}$, while the range of frequencies used for the slightly undulating interfaces was between $1/10$ m$^{-1}$ and $1/60$ m$^{-1}$. These values are smaller than those used by Vantassel et al. (2022a), making the soil-bedrock interface less variable in the present study. The bedrock V$_S$ was randomly varied following a uniform distribution between a lower bound of 360 m/s and an upper bound of 760 m/s. Lateral and vertical perturbations were imposed on both the soil and rock portions of the V$_S$ model to simulate inhomogeneities present in natural materials. The small-scale irregularities introduced by the perturbations were 1 m to 2 m in the vertical direction and 4 m to 6 m in the horizontal direction. Further details on the computations used to introduce these small-scale irregularities, including the assumed correlation structure, can be found in Vantassel et al. (2022a).

The 100,000 synthetic models were developed by randomly changing the stiff soil V$_S$ multiplier values, the weathered rock V$_S$, the V$_S$ lateral and vertical perturbations, the interface depth, and the interface undulation frequencies within the upper and lower bound for each variable. Following the development of each Vs image, a Vp image was generated by using a Poisson's ratio of 0.33 for soil and 0.2 for rock. The mass density image was constructed by assigning a value of 2000 kg/m$^3$ for soil and 2100 kg/m$^3$ for rock. Twenty randomly selected Vs images generated using the procedures described above are shown in Figure 4. Note that while the Vs images in Figure 4 have been selected at random, the number of images for each of the three model types (i.e., highly undulating, lightly undulating, and linear) have been selected to follow the distribution of model types in the training set (i.e., 30%, 60%, and 10%, respectively).

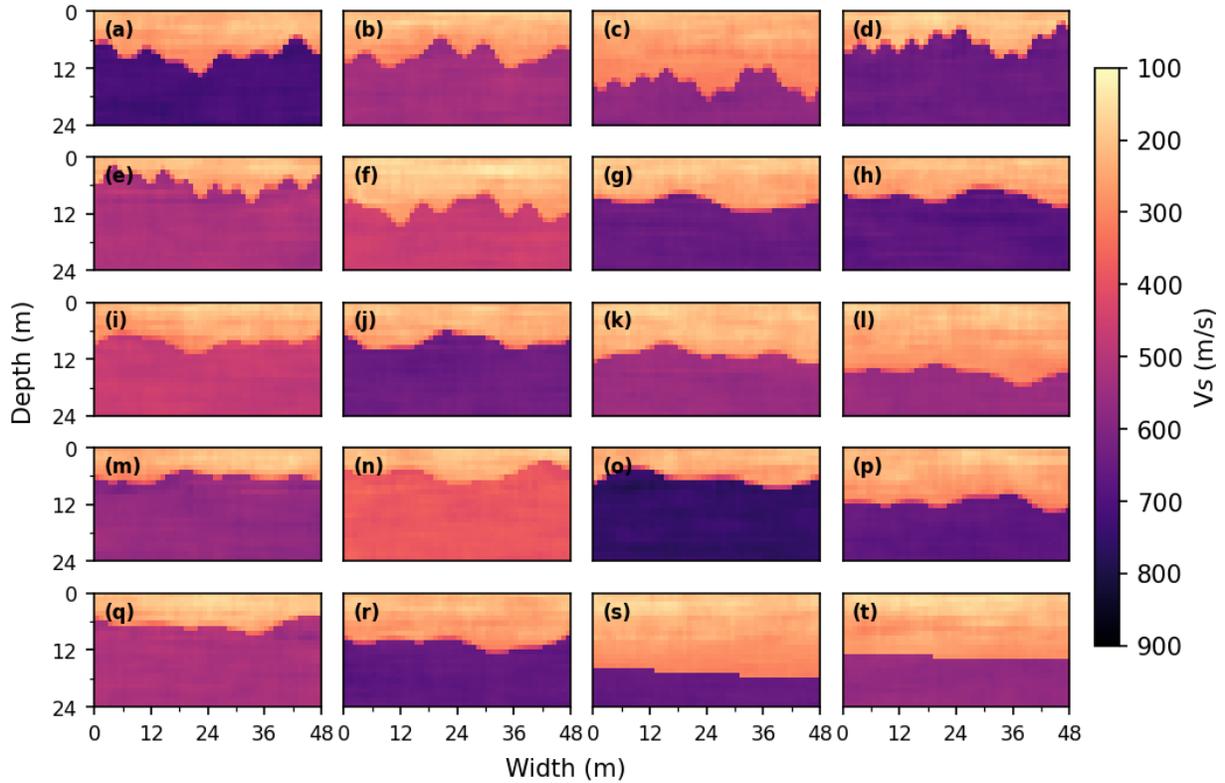

**Fig. 4.** Twenty randomly selected synthetic, near-surface 2D $V_S$ images from the 20,000 models used to test the convolutional neural networks (CNNs). The first six models (from a to f) have a highly undulating soil-rock interface, the following 12 models (from g to r) have a slightly undulating soil-rock interface, and the last two models (s and t) have a linear soil-rock interface.

## 4. Development of synthetic seismic wavefields

The 2D finite-difference software DENISE (Köhn, 2011; Köhn et al., 2012) was used to simulate elastic wave propagation through the synthetic models. Forty-eight receivers were placed at 1-m spacing across the center of the 104-m wide models to sample the wavefield generated by a source located 5 m to the left of the first receiver. This contrasts with the 24 receivers at a 2-m spacing and source located at the center of the array used by Vantassel et al. (2022a). The receivers occupied the distance between 28 m to 75 m, so they were as far away from the model boundaries as possible (refer to Figure 2a). A 30-Hz Ricker wavelet (Figure 3a) was used as a forcing function during the development of the training, validation, and testing datasets. This is a higher-frequency wavelet than that used by Vantassel et al. (2022a), allowing, in theory, high-resolution predictions to be developed. The finite-difference simulations were used to model two seconds of wave propagation to allow for recording of all wave types by the receivers. A sixth-order finite-difference operator in space and a second-order finite difference operator in time were utilized during the simulations. Perfectly matched layer absorbing boundaries (Komatitsch and Martin, 2007) were placed at the sides and the bottom of the domain, while the ground surface was modeled using a free boundary condition (Levander, 1988). Prior to wave propagation simulations, the soil models were interpolated from 1-m pixels to 0.2-m pixels to ensure that the wavefield was spatially sampled with a small enough grid to avoid numerical artifacts and instabilities, as recommended by Köhn et al. (2011). A 5E-5 second time step was used to satisfy the Courant-Friedrichs-Lewy criterion (Courant et al., 1967). The waveforms at each receiver location were recorded at a 400-Hz sampling rate. The simulations were performed on the Texas Advanced Computing Center's (TACCs) high-performance cluster Stampede2 using a single Skylake (SKX) compute node. Each simulation took approximately 5 seconds to complete.

*4.1 Post-processing to obtain the wavefield and dispersion input images*

To evaluate the relative performance between the time-distance and frequency-velocity CNNs, two identical datasets for training and testing of the two CNNs were developed. A total of 100,000 near-surface $V_S$ images and their corresponding seismic wavefields formed the image pairs in the dataset developed for the time-distance CNN. The same soil models and the dispersion images derived from their respective wavefields composed the dataset produced for the frequency-velocity CNN. Even though the synthetic models were 104-m wide and 24-m deep, only the 48 m at the center of the models immediately below the receivers were considered in training and testing the CNN, making the utilized soil models 48-m wide and 24-m deep, as illustrated in Figure 4. The near-surface $V_S$ images were then normalized by the maximum $V_S$ value in the training set to facilitate the training process. The waveforms recorded by the 48 receivers used for the time-distance CNN were normalized by the maximum amplitude across all receivers to preserve the rate of amplitude decay. The input shape for the time-distance CNN was 48x800x1, which represents the 48-receiver wavefield sampled for two seconds at 400 Hz.

The frequency-velocity CNN utilizes a normalized dispersion image as input. The dispersion image is computed using a wavefield transformation of the time-distance wavefield recorded by a linear array of receivers. Several wavefield transformation techniques can be used to generate a dispersion image from a recorded time-distance wavefield. For example, the frequency-wavenumber (f-k) (Gabriels et al., 1987; Nolet and Panza, 1976), slant-stack (McMechan and Yedlin, 1981), phase-shift (Park et al., 1998), and frequency-domain beamformer (FDBF) (Zywicki, 1999) transformations are all commonly used in various commercial and open-source surface wave processing software. The FDBF approach with a plane wave steering vector and no amplitude weighting (Zywicki, 1999) was used in the present study to develop the dispersion images used in training and testing the frequency-velocity CNN. We chose the FDBF over the more common frequency-wavenumber (f-k) approach because the FDBF technique allows the transformation of time-distance wavefields collected using non-uniformly spaced arrays of receivers and the calculation of dispersion power above the aliasing wavenumber (Vantassel and Cox, 2022a), thereby making the inputs to the CNN easier to acquire in the field. Furthermore, the FDBF approach has been judged to be superior to other wavefield transformation methods as assessed by Rahimi et al. (2021). Along with the FDBF, we use the plane wave steering vector to be consistent with the wavefield simulations. In particular, the seismic wavefield data were produced under the 2D, plane-strain assumption, and as a result are representative of waveforms generated by a line source (i.e., a plane wave source). Importantly, the FDBF method can be adjusted when applied to field data that is truly 3D in nature by swapping out the plane-wave steering vector and no amplitude weighting for cylindrical-wave steering and square-root-distance weighting to appropriately compensate for the effects of radiation damping in 3D data (Zywicki and Rix, 2005). This substitution removes the need to "correct" 3D waveforms collected in the field to 2D equivalents before dispersion processing and use in the frequency-velocity CNN.

The dispersion images in this study were generated using a frequency range of 5 Hz to 80 Hz with a 1-Hz frequency step and a phase velocity range of 100 m/s to 1000 m/s with 2.25 m/s velocity step. The dispersion image, therefore, had an input shape of 75x400x1 (i.e., 75 frequencies x 400 phase velocities). After computing the MASW dispersion images we utilized frequency-dependent normalization to further simplify the learning task of the CNN. With frequency-dependent normalization, sources with different frequency contents and offsets appear more similar than with other forms of normalization, such as absolute maximum normalization proposed by Park et al. (1998). By performing frequency-dependent normalization, we remove the need for the CNN to learn this aspect of the underlying physics, allowing for better generalization across various source types and locations. All dispersion images in this study were generated programmatically using the open-source Python package *swprocess* (Vantassel, 2021).

## 5. Time-distance and frequency-velocity CNN architectures

CNNs are an excellent tool for computer vision tasks and have shown great potential for use in seismic imaging (Wu and Lin, 2019; Yang and Ma, 2019; Liu et al., 2020; Vantassel et al., 2022a). The time-distance and frequency-velocity CNNs follow the architecture proposed by Vantassel et al. (2022a) of five convolutional layers interspersed with max-pooling layers. The convolutional layers employ a set of kernels, also called feature detectors or stencils, to capture the relevant patterns (i.e., feature maps) in the dataset images. Once the relevant features of the images are detected, subsampling or pooling layers are utilized to decrease the feature maps' spatial resolution, which in turn reduces the reliance on precise positioning within feature maps produced by the convolutional layers. Disregarding the exact position of features within a feature map while maintaining the relative position of features with respect to each other allows for a better CNN performance on inputs that differ from the training data. Max-pooling layers were used in the current study, as they were shown to be superior in capturing invariances in image-like data compared to subsampling layers (Scherer et al., 2010). The final convolutional layer is flattened and connected to a fully connected layer to perform the regression task. Table 1 shows the architectures used for the two CNNs. Google Collab and the open-source machine learning library Keras (Chollet et al., 2015) were used in training and testing the CNNs. 70%, 10%, and 20% of the developed 100,000 image pairs were used in the two CNNs training, validation, and testing stages, respectively. The two CNNs' architectures were adjusted from that originally proposed by Vantassel et al. (2022a) to accommodate their respective input sizes.

In addition to the networks' architectures, the model's hyperparameters need to be rigorously tuned to provide optimal performance. In the present study, we tuned the following hyperparameters by varying them between the upper and lower bounds listed below. We then selected the set that produced the best performance on the validation set. The hyperparameters considered include: the learning rate (0.1 to 0.0001), batch size (8 to 64), number of training epochs (10 to 100), optimizer (RMSprop and Adam), and loss function (mean squared error and mean absolute error). Ultimately, a learning rate of 0.0005, batch size of 16, training epoch of 40, Adam optimizer (Kingma and Ba, 2014), and mean absolute error (MAE) were selected for both the time-distance and frequency-velocity CNNs. We note these hyperparameters are similar to the ones selected by Vantassel et al. (2022a) despite being chosen after independent hyperparameter tuning exercises. The validation dataset MAEs for the time-distance and frequency-velocity CNNs using the selected hyperparameters are 0.022 and 0.025, respectively.

**Table 1.** Architectures for the time-distance and frequency-velocity convolutional neural networks (CNNs) developed in this study.

| Network layer type | Time-distance CNN | | Frequency-velocity CNN | |
|---|---|---|---|---|
| | Filter size | Size of output layer | Filter size | Size of output layer |
| 2D Convolution | 1x3 | 48x798x32 | 3x1 | 398x76x32 |
| 2D Max Pooling | 1x3 | 48x266x32 | 3x1 | 132x76x32 |
| 2D Convolution | 1x3 | 48x264x32 | 3x1 | 130x76x32 |
| 2D Max Pooling | 1x3 | 48x88x32 | 3x1 | 43x76x32 |
| 2D Convolution | 1x3 | 48x86x64 | 3x1 | 41x76x64 |
| 2D Max Pooling | 2x3 | 24x28x64 | 1x3 | 41x25x64 |
| 2D Convolution | 3x3 | 22x26x128 | 3x3 | 39x23x128 |
| 2D Max Pooling | 2x2 | 11x13x128 | 3x3 | 13x7x128 |
| 2D Convolution | 3x3 | 9x11x128 | 3x3 | 11x5x128 |
| Flatten | | 12672 | | 7040 |
| Dense | | 1152 | | 1152 |
| Reshape | | 24x48 | | 24x48 |

### 5.1. CNNs accuracy evaluation

The accuracy of the time-distance and frequency-velocity CNNs were evaluated using their respective 20,000 testing image pairs, which the networks were not trained on. The mean absolute percent error (MAPE) and MSSIM were used to provide a quantitative assessment of the CNN's performance. MAPE is the mean of the absolute value of the pixel-by-pixel percent error of each predicted $V_S$ image in physical units. The MSSIM index assess the quality of one image relative to another that is deemed to be of perfect quality based on three key features: luminance, contrast, and structure (Wang et al., 2004). MSSIM was calculated using a Gaussian windowing approach to match the implementation of Wang et al. (2004) and using a dynamic range equal to the difference between the absolute maximum and minimum $V_S$ values of the true models in the testing set.

The average values of MAPE and MSSIM for the time-distance CNN are 5.3% and 0.80, respectively, while the MAPE and MSSIM for the frequency-velocity CNN are 6.0% and 0.78, respectively (refer to Table 2). While the time-distance CNN provides slightly better accuracy than the frequency-velocity CNN, their performances are quite similar. Furthermore, we will show that the advantages permitted by operating in the frequency-velocity domain in terms of flexibility and generalization across data acquisition configurations outweigh the minor loss of accuracy. Figures 5 and 6 show the Vs image predictions for the time-distance and the frequency-velocity CNNs, respectively, for comparison with the 20 true images depicted in Figure 4. As noted above, Figure 4 depicts six highly undulating, 12 slightly undulating, and two linear soil-rock interface models. The MAPE and MSSIM values for each CNN prediction relative to the true image are also provided in Figures 5 and 6. While, on average, the time-distance CNN slightly outperforms the frequency-velocity CNN in terms of overall MAPE and MSSIM, there are some individual models for which the frequency-velocity CNN is slightly more effective (e.g., model S in Figures 5s and 6s). Furthermore, by-eye it would be difficult to distinguish the time-distance and frequency-velocity predictions from one another.

**Table 2.** The mean absolute percent error (MAPE) and mean structural similarity index (MSSIM) between the true $V_S$ images and the $V_S$ images predicted using the time-distance and frequency-velocity convolutional neural networks for the testing set and the different acquisition configurations. The values in the table represent the average MAPE and MSSIM over the number of images used for each acquisition configuration. The base acquisition configuration comprised 48 receivers with a 1-m spacing, a source offset of 5 m relative to the first receiver in the linear array, and a 30-Hz Ricker wavelet source forcing function.

| Acquisition variation | CNN used | Deviation from base configuration | Number of images | MAPE (%) | MSSIM |
|---|---|---|---|---|---|
| Base case | Time-distance | - | 20,000 | 5.3 | 0.80 |
| | Frequency-velocity | - | 20,000 | 6.0 | 0.78 |
| Varying receiver spacing | Frequency-velocity | 24 receivers at 2-m spacing | 5,000 | 6.0 | 0.77 |
| | | 16 receivers at 3-m spacing | 5,000 | 9.9 | 0.65 |
| | | 12 receivers at 4-m spacing | 5,000 | 24.0 | 0.32 |
| Varying source location | Frequency-velocity | Source at 6 m from first receiver | 5,000 | 6.8 | 0.77 |
| | | Source at 10 m from first receiver | 5,000 | 11.0 | 0.69 |
| | | Source at 20 m from first receiver | 5,000 | 12.7 | 0.64 |
| | | Average between 5-m source and 20-m source offsets | 5,000 | 8.1 | 0.73 |
| | | Average between 10-m source and 20-m source offsets | 5,000 | 11.2 | 0.67 |
| Varying source forcing function | Frequency-velocity | 15-Hz high-cut filtered spike forcing function | 5,000 | 5.9 | 0.78 |
| | | 3-Hz to 80-Hz linear sweep over 12-seconds forcing function | 4,887 | 7.3 | 0.75 |

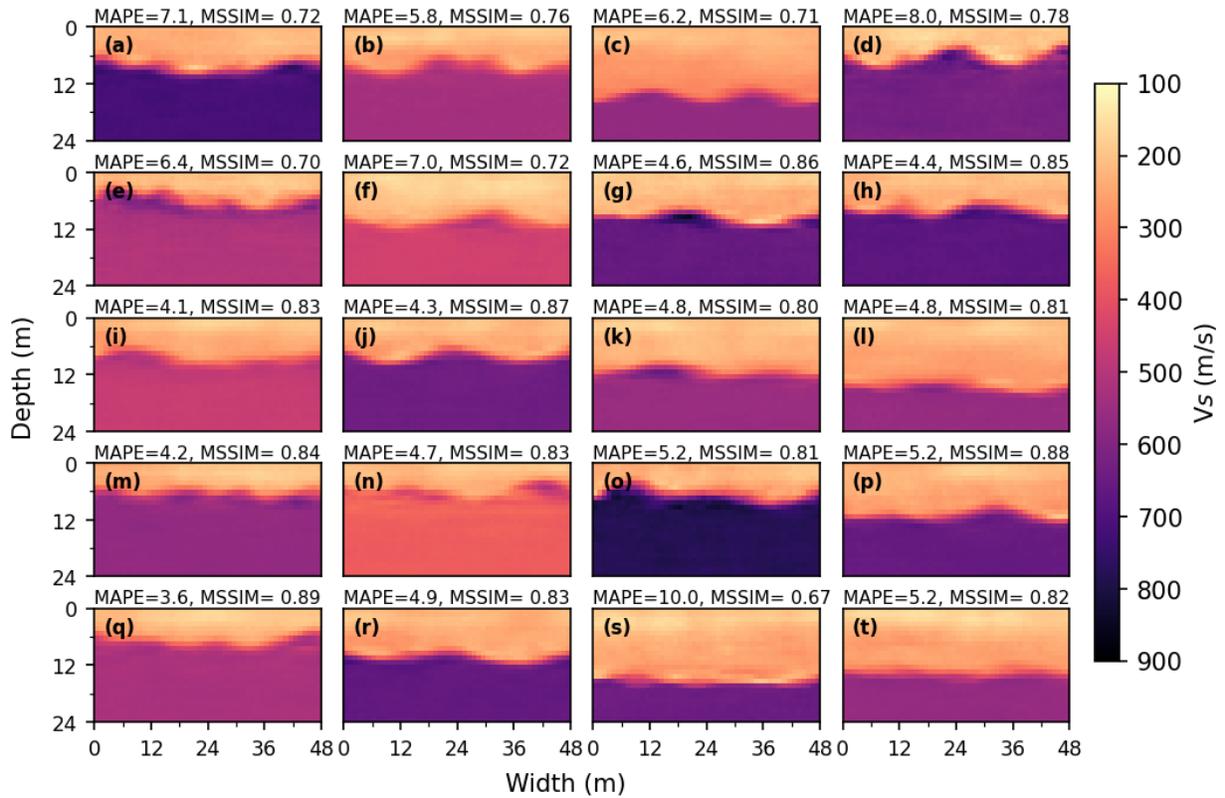

**Fig. 5.** The time-distance CNN's predictions of the true synthetic 2D Vs images presented in Fig. 4. The inputs used to obtain these predictions are the wavefields recorded by 48 receivers at 1-m spacings, which were excited by a 30-Hz Ricker source wavelet at 5 m from the first receiver. The mean absolute percent error (MAPE) and mean structural similarity index (MSSIM) of each predicted image are presented above each panel.

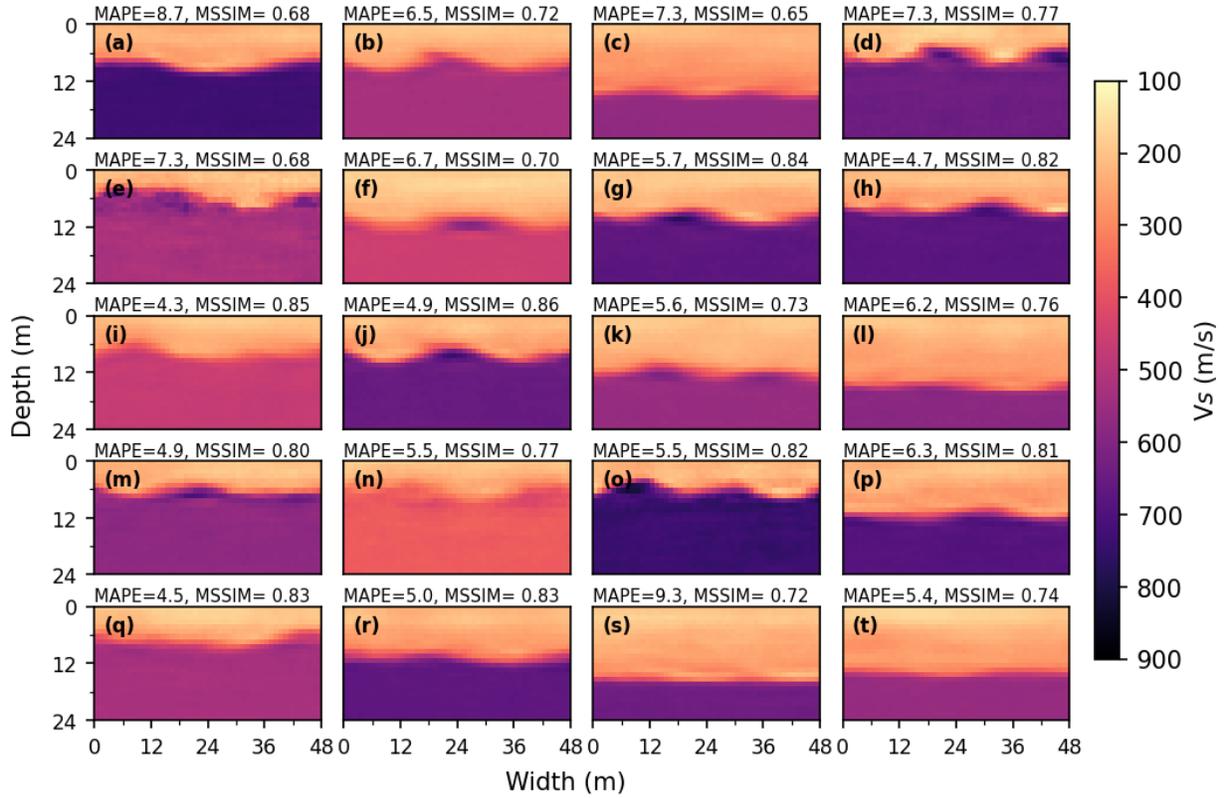

**Fig. 6.** The frequency-velocity CNN's predictions of the true synthetic 2D Vs images presented in Fig. 4. The inputs used to obtain these predictions are the normalized dispersion images obtained by post-processing the wavefields recorded by 48 receivers at 1-m spacings, which were excited by a 30-Hz Ricker source wavelet at 5 m from the first receiver. The mean absolute percent error (MAPE) and mean structural similarity index (MSSIM) of each predicted image are presented above each panel.

Figure 7 shows the residuals between the predicted frequency-velocity $V_S$ images (Figure 6) and the true images (Figure 4). It can be seen that large portions of the residual images have neutral colors, indicating relatively small differences in Vs between the true and predicted images. Specifically, the Vs of the soil and rock layers are generally well predicted, whereas most of the error is concentrated at the undulating soil-rock interfaces. While only shown herein for the frequency-velocity CNN, similar, localized interface errors are present in the time-distance predicted images and were also reported by Vantassel et al. (2022a) for their time-distance CNN. Nonetheless, the interface locations and major undulations are still fairly well preserved in the predictions.

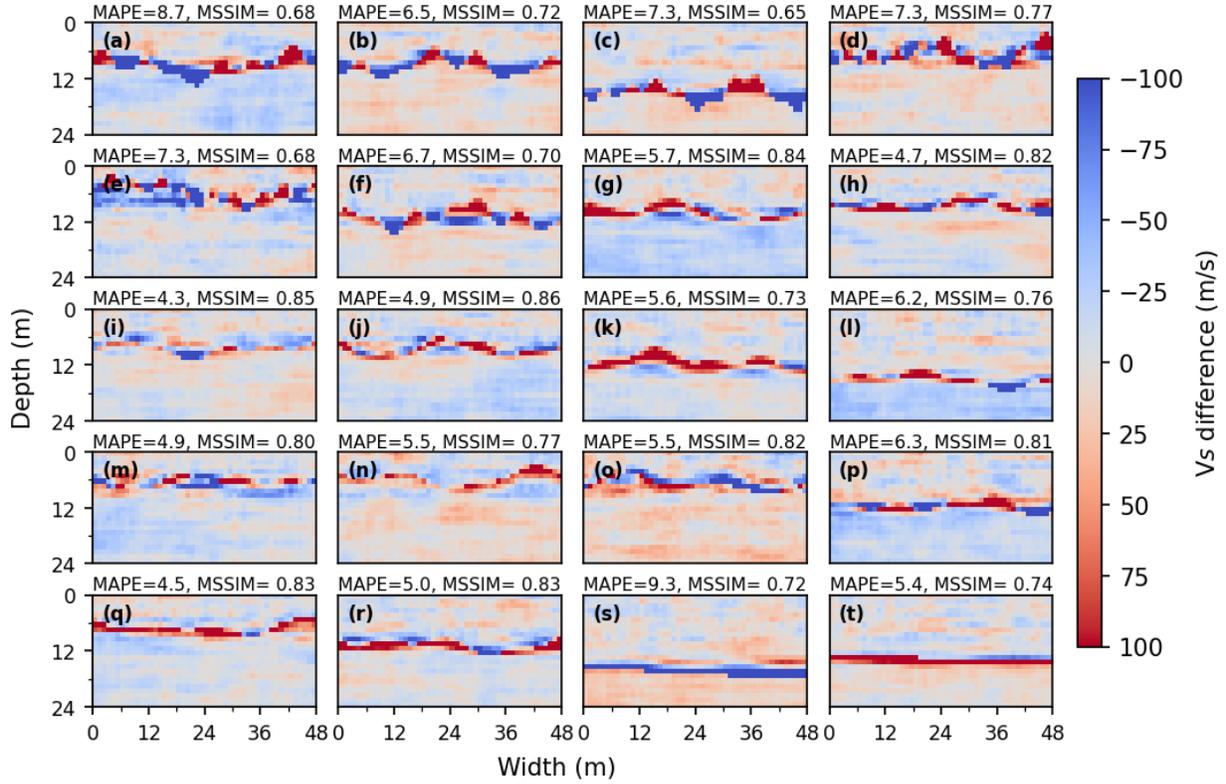

**Fig. 7.** The pixel-by-pixel difference between the twenty true 2D $V_S$ near-surface images presented in Fig. 4 and the corresponding frequency-velocity CNN image predictions shown in Fig. 6. The mean absolute percent error (MAPE) and mean structural similarity index (MSSIM) of each predicted image are presented above each panel.

## 6. Generalizing the frequency-velocity CNN across acquisition configurations

The acquisition generalization capabilities of the frequency-velocity CNN are evaluated by simulating different testing configurations than the base case configuration used during model training. The testing setup variations investigated herein include modifications to the number of receivers and receiver spacings, the source offset, and the source forcing function. The effects of each variation in the testing configuration are assessed separately, while the remainder of the setup is maintained identical to the base case configuration used in training the CNN. As a reminder, this base case training configuration was: 48 receivers with a 1-m spacing, a source offset of 5 m relative to the first receiver in the linear array, and a 30-Hz Ricker wavelet source forcing function (refer to Figure 2b). Due to the high computational costs required to run numerous wave propagation simulations with various acquisition configurations, only 5,000 of the 20,000 testing models were evaluated during this stage. These 5,000 models were randomly selected, but with the stipulation that this smaller population of models maintained the same ratios of 30% highly undulating interfaces, 60% slightly undulating interfaces, and 10% linear soil-rock interfaces as the original training and testing sets. We evaluate the performance of the frequency-velocity CNN image predictions for these 5,000 testing models with wavefields recorded using a wide range of simulated acquisition testing configurations using the same MAPE and MSSIM statistics used to evaluate the full 20,000 testing model set. The following sections discuss the performance of the frequency-velocity CNN for different testing configurations.

### 6.1. Generalizing to the number of receivers and receiver spacings

In the field, the number of receivers used to image the subsurface is dependent on equipment availability, testing space, and the objective of the experiment (e.g., better near-surface resolution versus greater imaging depth). Therefore, a CNN that can provide accurate 2D images from wavefields collected with different numbers of receivers is desirable. To test the frequency-velocity CNN for such acquisition generalization ability, three sets of 5,000 input dispersion images were obtained from wave propagation simulations on the 5,000 testing models. The first set of dispersion images was generated using the wavefields from 24 receivers at 2-m spacing, the second set from 16 receivers at 3-m spacing, and the third set from 12 receivers at 4-m spacing. The accuracy of the CNN's predictions for the three sets of inputs is presented in Table 2 in terms of MAPE and MSSIM. From Table 2, it is clear that the frequency-velocity CNN is capable of generalizing across various numbers of receivers and receiver spacings, but only within reasonable adjustments to the base configuration. For example, the MAPE and MSSIM values for 24 receivers at 2-m spacing are equal to 6.0 and 0.77, respectively, essentially equivalent to those from the base configuration. The MAPE and MSSIM values for 16 receivers at 3-m spacing are equal to 9.9 and 0.65, respectively, indicating a slight degradation in performance. However, the MAPE and MSSIM values for 12 receivers at 4-m spacing are equal to 24.0 and 0.32, respectively, which show a significant reduction in predictive capabilities as fewer receivers with larger receiver spacings are used to record the wavefield. Reasons for these observations are investigated further by considering plots presented in Figure 8.

Figure 8a shows the true $V_S$ image depicted in Figure 4j. Figures 8b and 8c show the input dispersion image and CNN output Vs image, respectively, for the 48-receiver base configuration. The input dispersion images and output Vs images for modified acquisition configurations are shown in Figures 8d and 8e, respectively, for 24 receivers and in Figures 8f and 8g, respectively, for 12 receivers. Also shown in all dispersion images are the high frequency spatial array resolution limits (sometimes called the f-k aliasing limits) for the respective receiver configurations, which are plotted as dashed white lines. The f-k aliasing limits represent the largest wavenumber (k), or equivalently the smallest wavelength ($\lambda$), that can be measured without concern for spatial aliasing. The spatial aliasing limit is a constant wavelength that is equal to two-times the receiver spacing (i.e., at least two measurements per spatial wavelength). Note that proper spatial sampling is analogous to proper time-domain sampling following the Nyquist sampling theorem. However, unlike in time-domain sampling the presence of contaminating short wavelength (i.e., high frequency) waves are less common due to material damping, and as a result clear dispersion data above the spatial aliasing limit can be used, albeit cautiously (Foti et al., 2018).

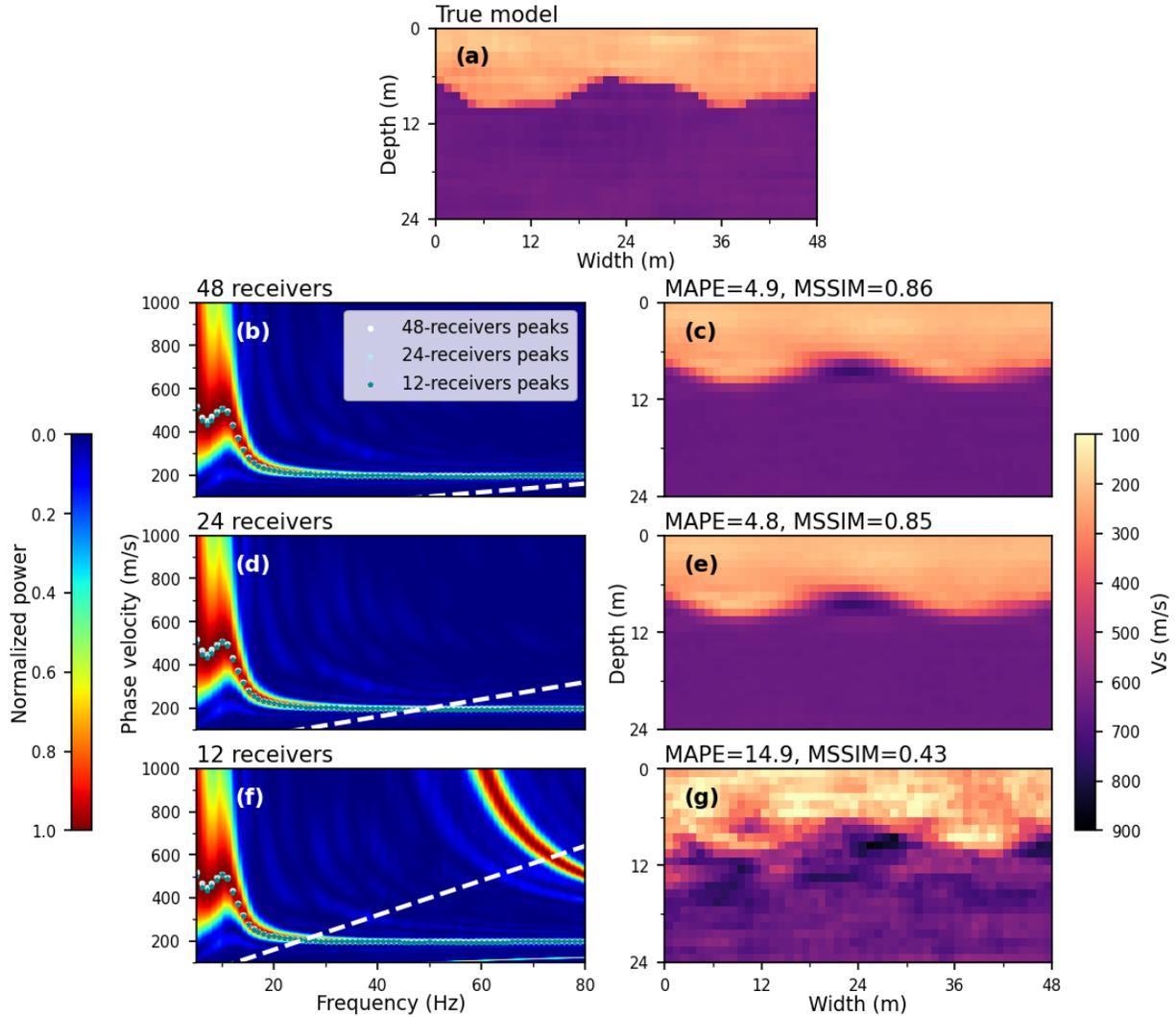

**Fig. 8.** (a) the true synthetic image from Fig. 4j. (b & c) the input dispersion image and output frequency-velocity CNN Vs image prediction, respectively, from the base receiver configuration (i.e., 48 receivers at 1-m spacing). (d & e) the input dispersion image and output frequency-velocity CNN Vs image prediction, respectively, from 24 receivers at 2-m spacing. (f & g) the input dispersion image and output frequency-velocity CNN Vs image prediction, respectively, from 12 receivers at 4-m spacing. The white dashed lines in (b), (d), and (f) represent the spatial array resolution limit for each array configuration. Peak power points obtained from the apparent fundamental Rayleigh wave mode of each dispersion image are also shown. The mean absolute percent error (MAPE) and mean structural similarity index (MSSIM) of each predicted image are presented above each panel.

As can be seen in Figures 8b, 8d, and 8f, the portion of the input dispersion image that exists below the spatial aliasing line increases as the number of receivers decreases and the receiver spacing increases. This means that when larger receiver spacings are used the higher frequency data may not be resolved accurately due to spatial aliasing. For example, the high-power trend at frequencies greater than 60-Hz in Figure 8f is not a true higher mode, but rather an artefact of spatial aliasing. While the frequency-velocity CNN can generalize across different receiver spacings to a certain extent, as evident by comparing the Vs-images MAPE and MSSIM values for the 48-receiver and 24-receiver configurations (refer to Figs. 8c and 8e), it cannot accurately generalize across receiver spacings that are drastically different from the base configuration, as the input dispersion images are affected by spatial aliasing. Even though the MAPE increases significantly when 12 receivers at 4-m increments are used (refer to Figure 8g), the CNN was still able to qualitatively predict the location of the soil rock interface quite well. Nonetheless, the actual Vs values for the soil and rock are not well resolved, a clear result of the limitations imposed by the CNN's training data (i.e., no spatial aliasing was present in the training set).

*6.2. Generalizing to source offset*

When performing linear-array, active-source surface wave testing, it is good practice to record the waveforms generated at several different source locations with increasing offset distance to help identify dispersion data contaminated by near-field effects and quantify dispersion uncertainty (Cox and Wood, 2011; Vantassel and Cox, 2022b). Near-field effects, which result from measuring low frequency (i.e., long wavelength) waves too close to the source location, are known to bias phase velocity estimates to lower velocities (Rosenblad and Li, 2011; Li and Rosenblad, 2011; Yoon and Rix 2009). As such, one needs to be cautious about placing an active source too close to a linear array. To complicate matters, near-field affects are site-dependent and, as a result, it is difficult to know *a-priori* what source offset distances are appropriate at a site. In addition to near-field effects, other wave propagation phenomena like body wave reflections and refractions influence the recorded seismic wavefield when varying source offset distances are used. This is particularly true when the subsurface conditions are neither 1D nor homogeneous. Therefore, a CNN that is capable of generalizing to different, or possibly multiple, source offsets is desirable.

To test the frequency-velocity CNN's ability to generalize in terms of source offset distance, five sets of 5,000 input dispersion images were obtained from wave propagation simulations on 5,000 testing models using different source offset distances relative to 48 receivers with 1-m spacing. Specifically, the first set of dispersion images was generated using a 6-m source offset, the second set using a 10-m offset, and the third set using at 20-m offset. The fourth and fifth sets of dispersion images were generated by combining dispersion images from two different source offsets in the frequency-phase velocity domain as follows: the fourth set resulting from stacking dispersion images from a 5-m and 20-m offset, and the fifth set from stacking dispersion images from a 10-m and 20-m offset. The accuracy of the CNN's predictions for the five sets of varied source offset inputs is presented in Table 2 in terms of MAPE and MSSIM. From Table 2, it is clear that, as with the receiver number and spacing discussed previously, the frequency-velocity CNN is capable of generalizing across various source offset distances, but only within reasonable adjustments to the base configuration. For example, the MAPE and MSSIM values for the 6-m source offset are only slightly worse than the base configuration, while the values for the 10-m offset and 20-m offset show increasing error as the source offset distance increases. Nevertheless, the average MAPE for the 20-m offset only increase by approximately 7% relative to the base configuration (i.e., 12.7% compared to 6.0%).

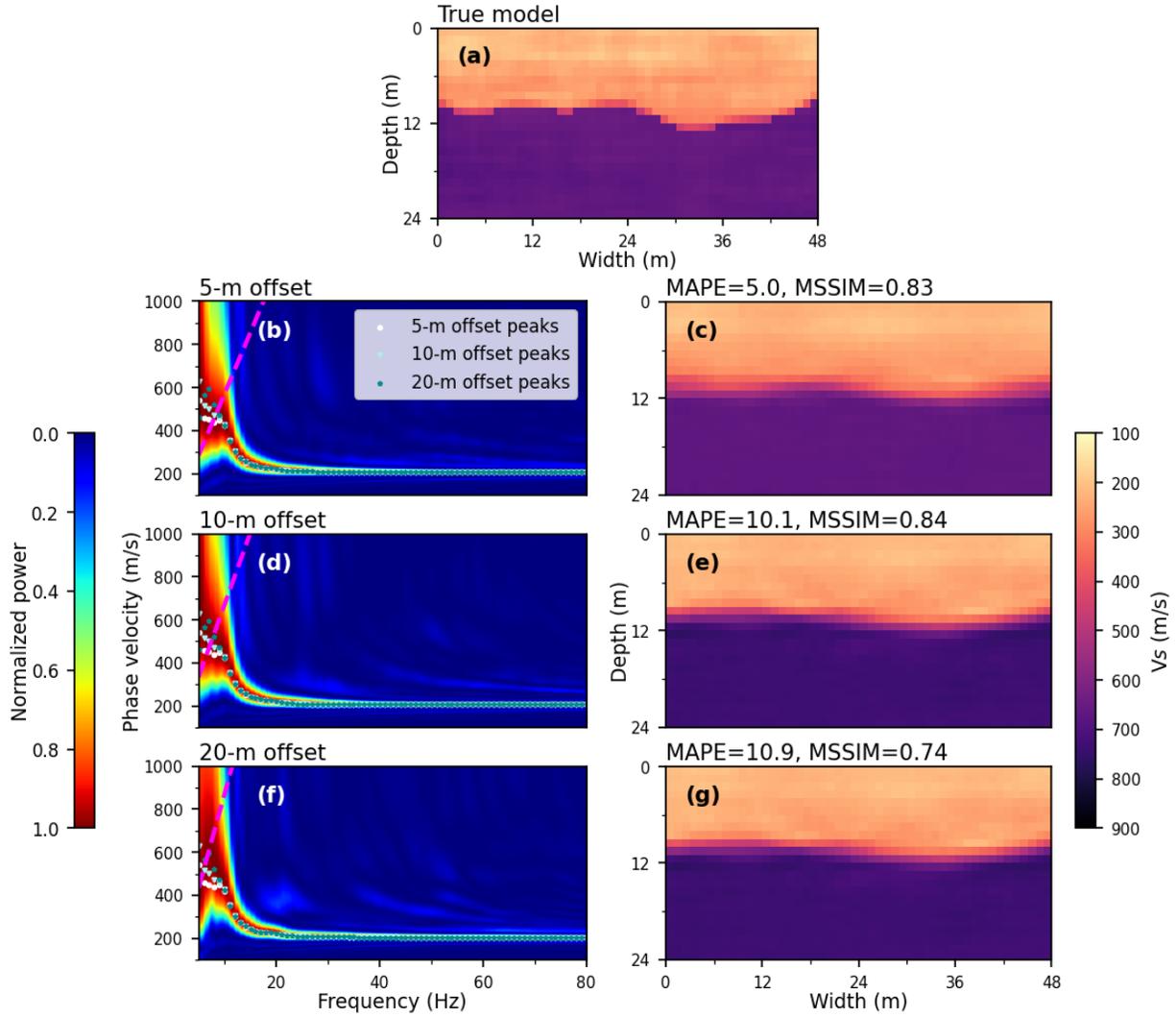

**Fig. 9.** (a) the true synthetic model from Fig. 4r. (b & c) the input dispersion image and output frequency-velocity CNN Vs image prediction, respectively, from the base receiver configuration (i.e., 5-m source offset). (d & e) the input dispersion image and output frequency-velocity CNN Vs image prediction, respectively, from a 10-m source offset. (f & g) the input dispersion image and output frequency-velocity CNN Vs image prediction, respectively, from a 20-m source offset. The magenta dashed lines represent constant wavelengths set equal to two-times the array-center distance that correspond to phase velocity errors less than 5% as presented by Yoon and Rix (2009). Peak power points obtained from the apparent fundamental Rayleigh wave mode of each dispersion image are also shown. The mean absolute percent error (MAPE) and mean structural similarity index (MSSIM) of each predicted image are presented above each panel.

Figure 9a shows the true V$_S$ image depicted in Figure 4r. Figures 9b and 9c show the input dispersion image and CNN output Vs image, respectively, for the 5-m source offset base configuration. The input dispersion images and output Vs images for modified acquisition configurations are shown in Figures 9d and 9e, respectively, for a source offset of 10-m and in Figures 9f and 9g, respectively, for a source offset of 20-m. Also shown in all dispersion images are magenta dashed lines that delineate zones where the low frequency (i.e., long wavelength) dispersion data from the various source offset distances may be influenced by near-field effects. These lines represent constant wavelengths set equal to two-times the array-center distances that correspond to phase velocity errors less than 5%, as presented by Yoon and Rix (2009). The array-center distance is equal to the distance from the source to the center of the linear array. Hence, the larger the source offset distance, the greater the array-center distance, and the longer the maximum wavelength that can be extracted from the dispersion data without contamination from near-field effects. As noted above, near-field effects manifest in dispersion data as phase velocity estimates that are lower than actual conditions. Near-field effects can be observed at low frequencies in the dispersion images shown in Figure 9 by comparing the peak power points from the 5-m, 10-m, and 20-m source offset dispersion images. The peak power points from the 20-m offset have the highest phase velocity values at low frequencies, while those from the 5-m offset have the lowest phase velocity values. These differences are only visible for frequency-phase velocity pairs near the magenta dashed lines that delineate zones where low frequency data may be influenced by near-field effects. Other than these low frequency zones, the dispersion images from the various source offsets appear very similar to one another. As such, the increasing errors resulting from increasing source offset are simply due to the fact that the CNN was trained on dispersion images generated from only a single source offset (i.e., 5 m) and this single source offset had dispersion data at lower frequencies that were slightly biased to lower velocities due to near-field effects. This does not imply that the dispersion images used for training the CNN are incorrect, but rather that the network learned to associate the near-field dispersion data with its corresponding true Vs image. We observe that the dispersion images obtained from larger source offsets have lesser near-field effects and, therefore, result in Vs image predictions that are stiffer at depth than the ones obtained from a 5-m offset. This is the greatest source of increasing error when larger source offset distances are used.

The dispersion images generated from several different offsets can be stacked together to balance out the impact of near-field effects and other differences in the wavefields caused by source location. Table 2 demonstrates that the increased error caused by using larger source offset distances can be combatted by stacking dispersion images from both near and far source offsets. This stacking of dispersion images is facilitated by normalizing each image by its absolute maximum power to counteract the varying dispersion image powers caused by the same source type being excited at different offset distances (i.e., closer sources having higher absolute dispersion power than distant sources). Once all images are normalized by their absolute maximum power, the images are summed and re-normalized by the maximum power at each frequency (i.e., frequency-dependent normalization). As discussed above regarding Table 2, this results in reduced MAPE values relative to using only a single, larger source offset. Vs images obtained from the frequency-velocity CNN after stacking the 5-m and 20-m source offset dispersion images are shown in Figure 10. These 20 predicted images can be compared to their ground truth images shown in Figure 4, time-distance CNN predictions in Figure 5, and the base case frequency-velocity CNN predictions in Figure 6. While the MAPE values associated with the predicted Vs images in Figure 10 are slightly greater than those in Figure 5 and Figure 6, the images, except for a few high and low velocity artifacts, do not look significantly different. In fact, as detailed in Table 2, the average MAPE values across all 5,000 testing models for the combined 5-m and 20-m source offsets are only about 2% higher than the average MAPE values for the base case configuration. This demonstrates the ability of the frequency-velocity CNN to generalize to a number of different source offset distances, enabling more flexible field data acquisition.

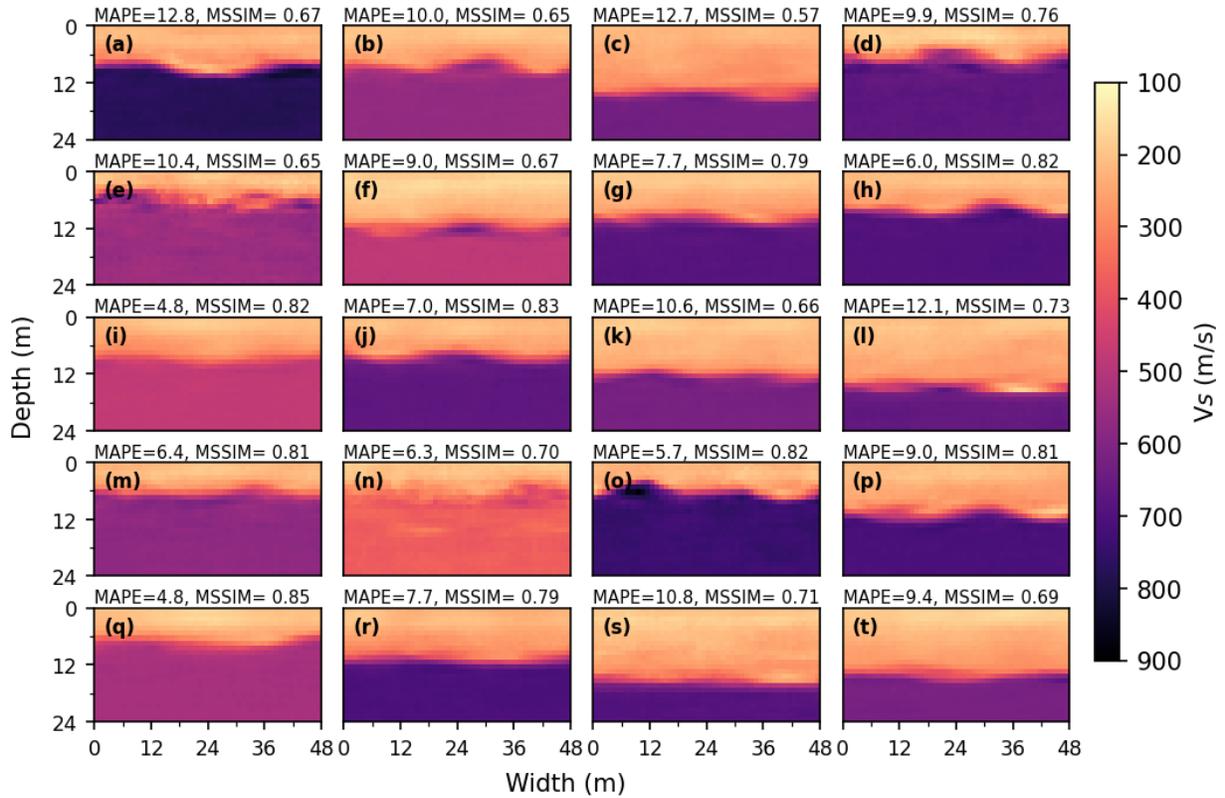

**Fig. 10.** The frequency-velocity CNN's predictions of the true synthetic 2D Vs images presented in Fig. 4. The inputs used to obtain these predictions are the normalized dispersion images obtained by averaging the 5-m and 20-m source offset dispersion images. The mean absolute percent error (MAPE) and mean structural similarity index (MSSIM) of each predicted image are presented above each panel.

### 6.3. Generalizing to source forcing function

Different source types are commonly used to excite the ground surface during linear-array, active-source wavefield testing. These active sources can range from large shaker trucks, to accelerated weight drops, to sledgehammer impacts, depending on the desired depth of profiling and the relative importance of the experiment. Therefore, a CNN that is capable of generalization in terms of providing accurate Vs images from wavefields collected with different source forcing functions is imperative for handling field applications.

To test the frequency-velocity CNN for acquisition generalization ability in terms of source forcing function, two sets of 5,000 input dispersion images were obtained from wave propagation simulations on 5,000 testing models using two different source forcing functions. Specifically, the first set of dispersion images was generated using a 15-Hz high-cut filtered spike wavelet, while the second set of dispersion images was generated using a 12-second-long linear chirp over frequencies from 3-Hz to 80-Hz. Recall that these two forcing functions, as well as the base case forcing function (i.e., a 30-Hz Ricker wavelet), are shown in both the time and frequency domains in Figure 3. It is clear from Figure 3 that the time and frequency domains of the source forcing functions are drastically different. However, we will demonstrate that the exact frequency-dependent amplitude of the source is not critical when using normalized dispersion images, provided the source induces broadband energy across the frequencies of interest.

The accuracy of the CNN's predictions for the two sets of varied source forcing functions is presented in Table 2 in terms of MAPE and MSSIM. From Table 2, it is clear that the frequency-velocity CNN is capable of generalizing across source forcing functions that are drastically different from one another. For example, the MAPE and MSSIM values for the 15-Hz high-cut filtered spike are virtually identical to those of the base configuration, which used a 30-Hz Ricker wavelet. Furthermore, the MAPE and MSSIM values for the 12-second chirp are only slightly worse than those for the base configuration. Note that the finite difference wave propagation simulations had to be extended to 13 seconds to capture the reflected and refracted waves from the extended chirp propagating through the models, and a total of 113 models out of the 5,000 testing models had to be discarded due to significant numerical artifacts caused by the length and complexity of the simulations. Despite these challenges, the average MAPE based on the 12-second chirp was less than 1.5% greater than the base configuration.

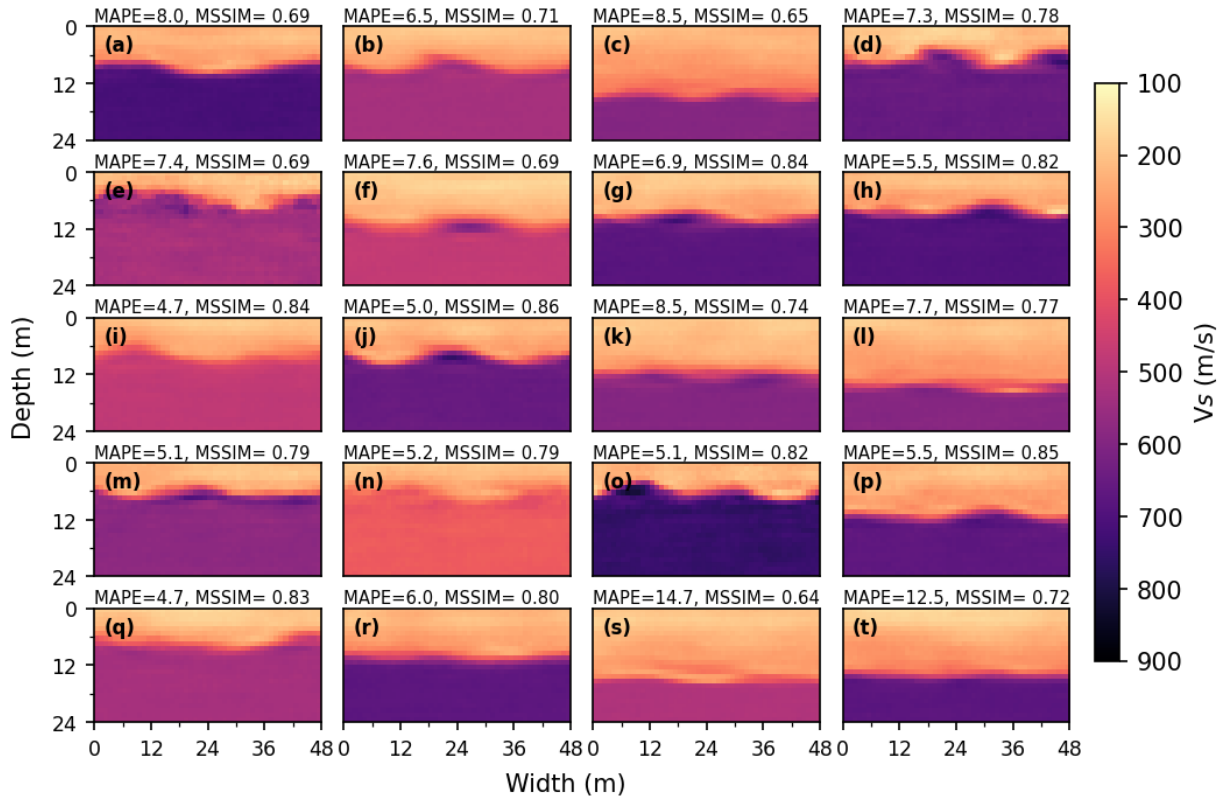

**Fig. 11.** The frequency-velocity CNN's predictions of the true synthetic 2D Vs images presented in Fig. 4. The inputs used to obtain these predictions are the dispersion images obtained by post-processing the waveforms recorded by 48 receivers at 1-m spacings, which were excited by a 12-seconds long 3-Hz to-80 Hz sweep/chirp at 5 m from the first receiver. The mean absolute percent error (MAPE) and mean structural similarity index (MSSIM) of each predicted image are presented above each panel.

The effects of using a 12-second chirp as a forcing function are more clearly visualized in Figure 11 by observing the Vs image predictions obtained from the frequency-velocity CNN. These 20 predicted images are for the same 20 true synthetic models shown in Figure 4, and the results can be compared directly to those shown for the time-distance CNN predictions in Figure 5 and the base case frequency-velocity CNN predictions in Figure 6. On average, the MAPE values associated with the predicted Vs images in Figure 11 are higher than those in Figures 5 and 6, but the differences are minimal, and the images appear to be quite similar. This exhibits notable generalization capabilities for the frequency-velocity CNN in terms of using different source forcing functions, provided they contain energy across the CNN input frequency band.

## 7. Field application and validation

The proposed frequency-velocity CNN was used to predict the near-surface 2D $V_S$ image at the NHERI@UTexas Hornsby Bend site in Austin, Texas, USA, where extensive site characterization studies have been conducted in recent years (e.g., Stokoe et al., 2020; Vantassel et al., 2022b). The wavefields used for testing the frequency-velocity CNN were actively generated using a sledgehammer to strike vertically on a square aluminum strike-plate. The wavefields were recorded by 24, 4.5-Hz vertical geophones. Five distinct sledgehammer blows were recorded at a distance of 5 m relative to the first geophone for subsequent stacking in the time domain to increase the signal-to-noise ratio. The input to the CNN was the dispersion image obtained from the FDBF with cylindrical-steering vector and square-root-distance weighting. Figure 12 shows the stacked waveforms, input dispersion image, and the predicted frequency-velocity CNN 2D $V_S$ image. The predicted Vs image indicates stiff soil (Vs ~ 200 to 300 m/s) overlying gently dipping rock, with an interface at ~ 12 to 14 m.

Two boreholes were recently drilled at the Hornsby Bend site to investigate subsurface layering down to rock; the first (i.e., B1) was located 12.5 m from the start of the geophone array, while the second (i.e., B2) was located 137.5 m away. Both borehole logs indicated a shale layer at approximately 13.5 m below the ground surface. Figure 12c shows a schematic of the lithology and 1D layer boundaries obtained from borehole log B1 superimposed at its correct location on the CNN-predicted 2D $V_S$ image. Based on Figure 12c, the CNN was not only capable of precisely determining the depth of the shale layer, but it was also capable of characterizing the increase in stiffness from the near-surface sandy silty clay (CL-ML) soils to the underlying clayey sand (SC) soils. This is an interesting finding since the frequency-velocity CNN was only trained on two-layer synthetic models (i.e., variable soil overlying undulating rock). Nonetheless, the 2D Vs image predicted by the frequency-velocity CNN appears to properly capture the expected trends in Vs for this three-layer field site. This is a particularly notable finding given that the boreholes at the site were drilled after the CNN predictions were developed, such that the 2D Vs image was produced in a truly blind manner without any *a priori* constraints from boring logs. It is worth mentioning that the CNN input dispersion image in Figure 12b was generated in less than one minute, while the 2D Vs prediction in Figure 12c took less than two seconds to obtain.

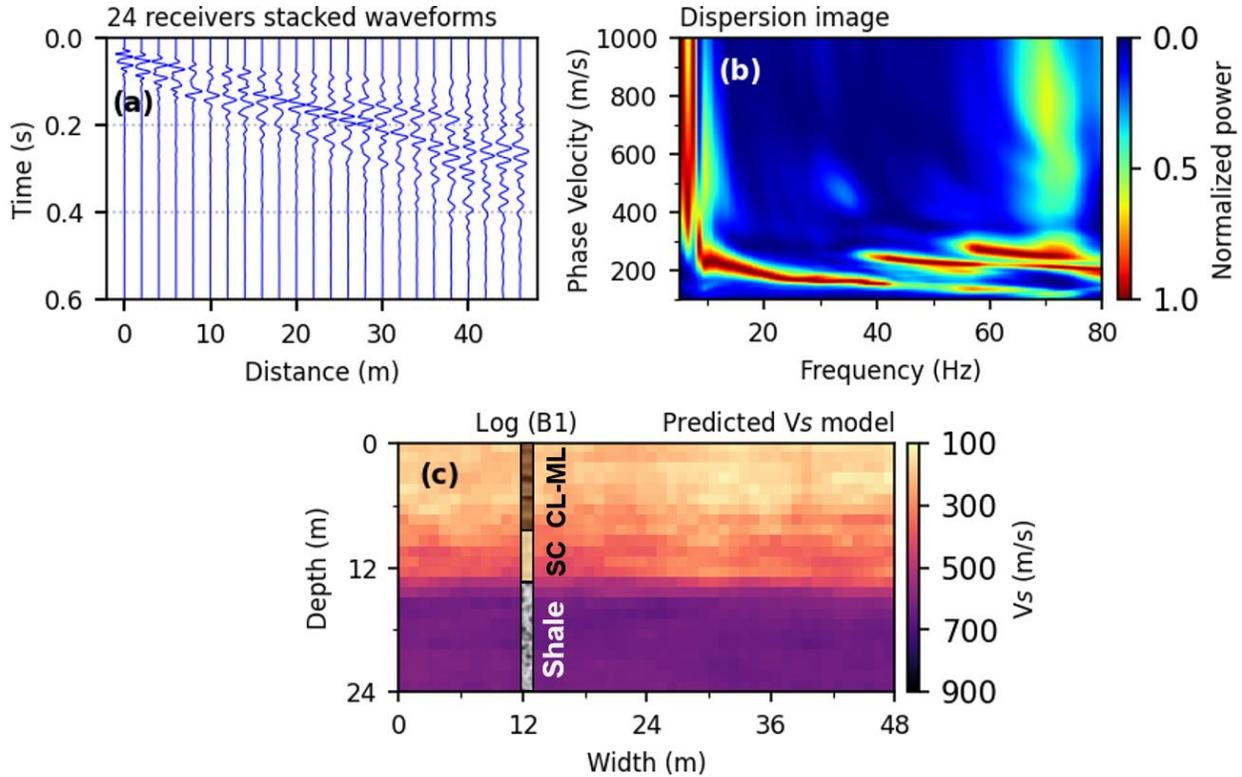

**Fig. 12.** Application of the frequency-velocity CNN to linear-array, active-source wavefield measurements collected at the Hornsby Bend site. (a) the staked waveforms from five sledgehammer impacts at 5-m source offset relative to the first geophone in a linear array of 24 receivers at 2-m spacing. (b) the dispersion image associated with the wavefield presented in (a), which was used as input for the frequency-velocity CNN. (c) the frequency-velocity CNN output 2D $V_S$ image for the Hornsby Bend site. For comparison with actual field conditions, a borehole log (i.e., B1) is superimposed on the predicted $V_S$ image at 12.5 m, which is the location where the boring was conducted.

While the borehole lithology log from B1 provides great insights into the 1D subsurface layering, it has not yet been used to perform downhole $V_S$ profiling and can therefore not be sued to judge the accuracy of the $V_S$ predictions. So, to compare the CNN predicted 2D $V_S$ image with other $V_S$ estimates at the Hornsby Bend site, the extensive 1D surface wave inversions performed by Vantassel et al. (2022c) were used. Vantassel et al. (2022c) processed MASW data collected at the Hornsby Bend site and inverted it using four layering parameterizations of 3, 5, 7, and 9 layers, based on the layering by number (LN) approach, to investigate subsurface layering uncertainty (Vantassel and Cox, 2021). They reported a suite of 1D Vs profiles representing the uncertainty in Vs across the MASW array. The median 1D Vs profiles obtained from each of these four layering parameterizations, along with the lognormal discretized median $V_S$ profile from all the inversion realizations investigated by Vantassel et al. (2022c), are provided in Figure 13. The entire MASW inversion process performed by Vantassel et al. (2022c) took approximately six hours to complete on four SKX nodes on the Stampede2 supercomputer cluster. Vantassel et al. (2022c) also reported three $V_S$ profiles along the array that were obtained from correlations to cone penetration testing (CPT) data, which are also shown in Figure 13. Each CPT-based 1D Vs profile was obtained from a CPT sounding by averaging the correlated $V_S$ values from three CPT-Vs relationships developed by Hegazy and Mayne (2006), Andrus et al. (2007), and Robertson (2009). To facilitate comparison between the CNN-predicted 2D $V_S$ image and the 1D Vs profiles reported by Vantassel et al. (2022c), the predicted 48-m wide 2D $V_S$ image was discretized into 48 1D $V_S$ profiles by slicing vertically through the 2D Vs image at 1-m increments, as shown in Figure 13. As can be seen in Figure 13, the median trend from the 1D MASW inversions reported by Vantassel et al. (2022c) is slightly stiffer (higher Vs) than the median trend from the 1D slices through the 2D Vs image obtained from the CNN prediction over the top 13 m, and slightly softer at depths greater than 13 m. Nevertheless, the agreement between the CNN and MASW Vs predictions is quite good. Figure 13 also shows good agreement between the 1D Vs profiles from the CNN prediction and the 1D $V_S$ profiles obtained from the CPT-Vs correlations reported by Vantassel et al. (2022c). Log B1 is also repeated in Figure 13 so that the expected soil-type lithology can be visualized relative to the 1D Vs profiles. The increasing stiffness from the CL-ML soils to the underlying SC soils is further validated by observing the trends in the standard penetration test (SPT) raw blow count (N) values, which are presented next to the B1 lithology log.

The strong agreement between the CNN predictions and the 1D Vs profiles derived from both MASW and CPT-Vs correlations, as well as the precision in locating the rock depth relative to the ground truth (i.e., log B1), demonstrate that the frequency-velocity CNN is capable of rapidly generating accurate Vs images for geologic conditions similar to those on which it was trained (i.e., stiff soil overlying rock). Furthermore, even though the network was trained using a 30 Hz Ricker wavelet source recorded by 48 receivers with a 1-m spacing, it was capable of utilizing field data collected with sledgehammer impacts recorded by 24 receivers with a 2-m spacing. This illustrates that the proposed frequency-velocity CNN can generalize across different field data acquisition configurations.

To further assess the 2D Vs image predicted by the frequency-velocity CNN, Figure 14 compares the measured and predicted MASW dispersion images. In particular, Figure 14a shows the measured MASW dispersion image that was computed from the stacked experimental waveforms from the sledgehammer source at the Hornsby Bend site. However, the predicted MASW dispersion image shown in Figure 14b required some extra work to obtain. First, the frequency-velocity CNN's predicted 2D Vs image (recall Fig. 12c) was used to obtain a full 2D predicted subsurface model by applying the simple $V_P$ and mass density rules discussed above in regards to synthetic model development. Then, finite difference wave propagation simulations were performed using a 30 Hz Ricker wavelet located 5 m from the array. The waveforms were measured on a 24-receiver array at a 2-m spacing to be consistent with the actual field data. Then, the dispersion image was obtained using the FDBF with plane-wave steering vector and no amplitude weighting. For the purpose of comparison, the peak dispersion data amplitudes in Figure 14a are repeated in Figure 14b. Overall, we observe consistency between the modal trends in the measured and predicted MASW dispersion image, with the R0 mode between 10 and 40 Hz showing particularly strong agreement. However, the higher modes observed in Figure 14a are not as pronounced in Figure 14b. Nonetheless, on careful inspection of Figure 14b it is possible to observe some higher mode trends that show some consistency with the locations of the higher mode dispersion data present in the experimental wavefield. This comparison serves to emphasize that, while not perfect, the 2D Vs image predicted by the frequency-velocity CNN (i.e., Figure 12b) is relatively consistent with the experimental wavefield data and measured dispersion image (i.e., Figure 12a) from the field experiment.

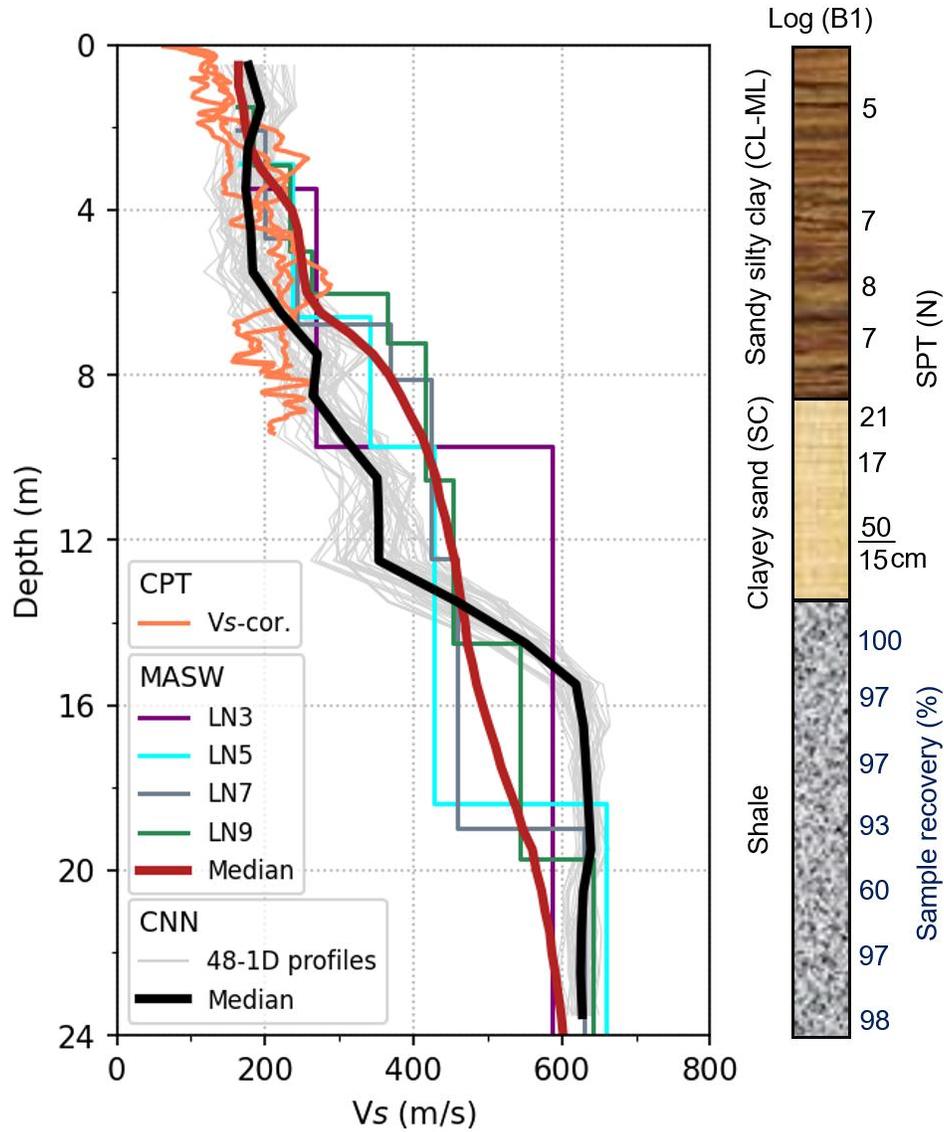

**Fig. 13.** Comparison of 1D Vs profiles for the area imaged by the frequency-velocity CNN at the Hornsby Bend site. The figure shows four layer-by-layer median 1D V$_S$ profiles from four layering by number (LN) MASW inversion parameterizations, as well as the overall lognormal discretized median 1D V$_S$ profile from the 1D MASW inversions performed by Vantassel et al. (2022c). The figure also shows three 1D V$_S$ profiles obtained from correlations with three CPT soundings, as reported by Vantassel et al. (2022). These 1D V$_S$ profiles are plotted relative to the 48 1D Vs profiles extracted from the 2D Vs image obtained from the frequency-velocity CNN and their lognormal median. To compare the CNN predictions with ground truth, the lithology log from borehole B1 is also provided.

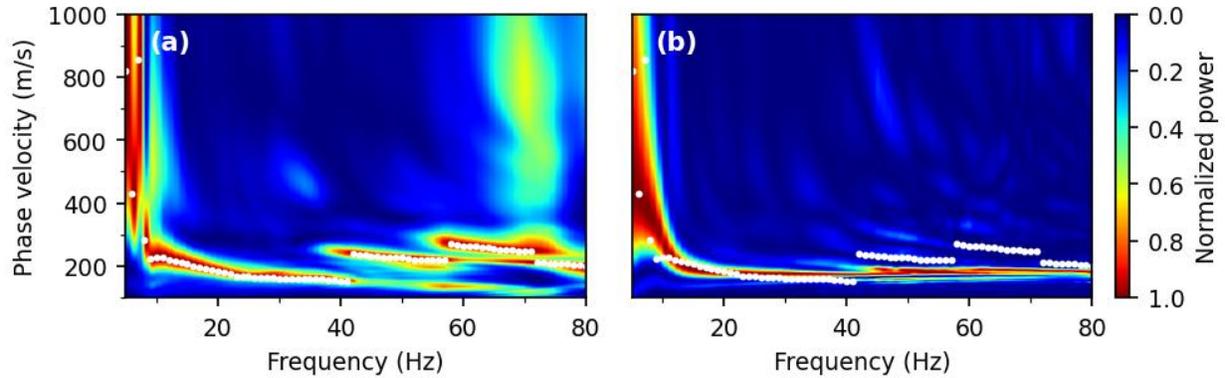

**Fig. 14.** (a) dispersion image developed using field measurements, and (b) the dispersion image obtained from the predicted 2D Vs model. To allow comparisons between the dispersion images, the peak power at each frequency in panel (a) is shown in panel (b) using white circles.

## 8. Conclusions

A frequency-velocity CNN has been developed for rapid, non-invasive 2D Vs near-surface imaging using stress waves. The CNN uses a normalized dispersion image as an input and outputs a 2D Vs image. The proposed framework provides significant flexibility in the linear-array, active-source experimental testing configuration used in generating the CNN input at a given site, accommodating various source types, source offsets, numbers of receivers, and receiver spacings. Such acquisition flexibility permits the use of the developed CNN as an end-to-end imaging technique, or as a means for generating rapid starting models for FWI. A total of 100,000 soil-over-rock synthetic models were used to train, validate, and test the CNN. The testing metrics of the developed frequency-velocity CNN revealed similar prediction accuracy to the time-distance CNN recently developed within our research group, which showed great promise but lacked flexibility important for field applications. The acquisition generalization ability of the proposed frequency-velocity CNN was first demonstrated using sets of 5,000 synthetic near-surface models. For each set of 5,000 models, the inputs to the CNN were dispersion images obtained using different testing configurations than the ones used during training the CNN. The CNN showed remarkable acquisition generalization ability with regards to the number of receivers, receiver spacings, source offset distances, and source forcing functions, as long as the testing configuration was not drastically different relative to the base case configuration on which the CNN was trained. Finally, the ability of the proposed CNN to handle field data was demonstrated using the experimental tests conducted at the Hornsby-Bend site in Austin, Texas, USA. The good agreement between the CNN's predicted 2D Vs image and the actual subsurface structure determined through 1D surface wave inversions, CPT-Vs correlations, and boring logs reinforce the capabilities of the proposed CNN for accurately retrieving 2DVs images using field data from testing configurations different from the one used during training.

### Acknowledgments

The open-source software DENISE (Köhn 2011; Köhn et al., 2012) was used for all the wave propagation simulations conducted in this study. The Texas Advanced Computing Center's (TACC's) cluster Stampede2 was used in the construction of seismic wavefield-image pairs, with an allocation provided by DesignSafe-CI (Rathje et al. 2017). Google Collab along with the open-source machine learning library Keras (Chollet et al., 2015) were used in training and testing the CNNs presented herein. The wavefield transformations were performed using the open-source Python package *swprocess* (Vantassel 2021). Matplotlib 3.1.2 (Hunter 2007) was used to create the figures in this study. This work was supported primarily by the U.S. National Science Foundation (NSF) grant CMMI-1931162 with equipment resources for field testing at the Hornsby Bend site associated with grant CMMI-2037900. However, any opinions, findings, and conclusions or recommendations expressed in this material are those of the authors and do not necessarily reflect the views of NSF.